%% file: main.tex
\documentclass{article} %
\PassOptionsToPackage{sort&compress}{natbib} %
\def\arxivversion{1}
\def\arxivheader{Preprint. Under review.}
\def\arxivpdfauthor{Zhibin Qin, Zhenxiong Tan, Xinchao Wang}
\def\arxivauthors{Zhibin Qin \qquad Zhenxiong Tan \qquad Xinchao Wang\thanks{Corresponding author.} \\
National University of Singapore \\
\texttt{e1352224@u.nus.edu, xinchao@nus.edu.sg}}
\ifdefined\arxivversion\else\documentclass{article}\fi
\usepackage{iclr2027_conference,times}

\input{math_commands.tex}

\usepackage{hyperref}
\usepackage{url}
\usepackage{graphicx}
\usepackage{booktabs}
\usepackage{tabularx}
\usepackage{amsmath,amssymb}
\usepackage{tikz}
\usepackage{colortbl}
\usepackage{algorithm}
\usepackage{algpseudocode}
\usetikzlibrary{arrows.meta}
\usepackage{caption}

\usepackage{xspace}

\newcommand{\methodname}{\textsc{FAVOR}\xspace}
\newcommand{\method}{\methodname}  %

\ifdefined\arxivversion
  \newcommand{\citationrev}[1]{#1}
\else
  \newcommand{\citationrev}[1]{\textcolor{red}{#1}}
\fi
\ifdefined\arxivversion
  \usepackage{placeins}                      %
  \definecolor{favorlink}{rgb}{0.21,0.49,0.74}
  \hypersetup{colorlinks=true, linkcolor=favorlink, citecolor=favorlink, urlcolor=favorlink,
    pdftitle={Future Anchored Verification and Online Recovery for World Action Models}}
  \ifdefined\arxivpdfauthor\hypersetup{pdfauthor={\arxivpdfauthor}}\fi
  \setcitestyle{numbers,square,citesep={,},aysep={,},yysep={,}} %
\fi

\title{Future Anchored Verification and \\ Online Recovery for World Action Models}

\ifdefined\arxivversion
  \author{\arxivauthors}
  \iclrfinalcopy %
\else
  \author{Anonymous Author(s) \\
  Affiliation \\
  Address \\
  \texttt{email}
  }
\fi

\ifdefined\arxivversion\else\fi
\begin{document}

\ifdefined\arxivversion\setcounter{footnote}{1}\fi %
\maketitle
\ifdefined\arxivversion\setcounter{footnote}{0}\fi %
\ifdefined\arxivheader\lhead{\arxivheader}\fi

\input{section/00_abstract}
\input{section/01_intro}
\input{section/02_related}
\input{section/03_Preliminary}
\input{section/04_method}
\input{section/05_experiments}

\input{section/07_conclusion}

\input{section/07_statements}

\clearpage %
\bibliography{iclr2026_conference,analysis_references}
\ifdefined\arxivversion\bibliographystyle{iclr2027_conference_nourl}\else\bibliographystyle{iclr2027_conference}\fi

\clearpage %
\appendix
\input{section/08_appendix}

\end{document}

%% file: math_commands.tex
\usepackage{amsmath,amsfonts,bm}

\def\eqref#1{equation~\ref{#1}}

\def\1{\bm{1}}

\DeclareMathAlphabet{\mathsfit}{\encodingdefault}{\sfdefault}{m}{sl}
\SetMathAlphabet{\mathsfit}{bold}{\encodingdefault}{\sfdefault}{bx}{n}



%% file: section/00_abstract.tex
\begin{abstract}
World action models (WAMs) have emerged as a promising paradigm for robotic manipulation. They act by first predicting how a task should be performed and then decoding the actions from that future. However, the remaining actions are invalid once execution drifts from the prediction. Simply replanning from the already out of distribution state rarely restores what the task still requires; existing execution monitors decide when to stop, but not what to restore. We observe that the answer is already in hand: the future the WAM predicted before acting depicts exactly the states it intended to pass through. We introduce \method{} (\textbf{F}uture \textbf{A}nchored \textbf{V}erification and \textbf{O}nline \textbf{R}ecovery), a lightweight framework that keeps these predicted frames as anchors and uses them for verification and recovery. An Anchor Verifier compares each observation with its anchor, together with the executed actions, to flag deviations that break the task. 
Anchor-Guided Recovery uses a vision-language model to turn the flagged anchor into a short corrective instruction. Under strengthened instruction guidance, the WAM executes this instruction to return to the intended future. It then resumes the task.
\method{} raises the task success of the base WAM from 97.85\% to 98.10\% on LIBERO and from 72.60\% to 72.98\% on LIBERO-Plus without modifying the policy.
\end{abstract}

%% file: section/01_intro.tex
\section{Introduction}
\label{sec:intro}

World action models (WAMs) have emerged as a promising paradigm for robotic
manipulation. They work in a forward and backward way: they first predict the visual future a task should pass through and then decode the actions that realize the future \citep{du2023unipi,tian2025seer,wu2024gr1,zhu2025uwm,ye2026dreamzero,yuan2026fastwam,li2026lingbotva,kim2026cosmos,ye2026gigaworld,zhou2026tau0wm,wam2026tutorial}. Unlike policies that map current observations and task instruction directly to actions \citep{chi2023diffusionpolicy,kim2024openvla,black2024pi0}, every action is thus grounded in an explicit future of where the scene should go.

That future is only a plan. Imprecise actions or external disturbances can push the scene off the predicted path, invalidating the remaining actions \citep{li2026badwam,wang2026trustimagination}. Simply replanning from the deviated state, however, offers little help: WAMs are trained on successful demonstrations \citep{ross2011dagger,laskey2017dart,reichlin2022back}, so neither the future predicted from such an unfamiliar state nor the actions decoded from it can be trusted. Execution monitors for WAMs help by flagging when the remaining actions should no longer be trusted \citep{wang2026trustimagination,liu2026checkvla,zhang2026foresight,agia2025sentinel,xu2025faildetect}, but say nothing about what to restore. Recovery methods address this gap, but obtain their target from outside the policy \citep{dai2024racer,pan2026vlacorrector}. This raises a natural question: can a WAM tell when execution has left its plan, and find its way back on its own?

Our observation is that the target of recovery is already in the policy's hands. The future the WAM predicted depicts the states it intended to pass through, and the actions were generated under its guidance, so the actions remain valid exactly as long as the scene follows it. When a deviation occurs, this prediction still shows the condition that was lost, such as a mug held in the gripper, and that state lies in the region where the policy was trained. A deviation therefore invalidates the remaining actions, but not the states they were meant to establish. This suggests a simple principle: use the policy’s own predicted future as an anchor.

\begin{figure}[t]
\centering
\includegraphics[width=\textwidth]{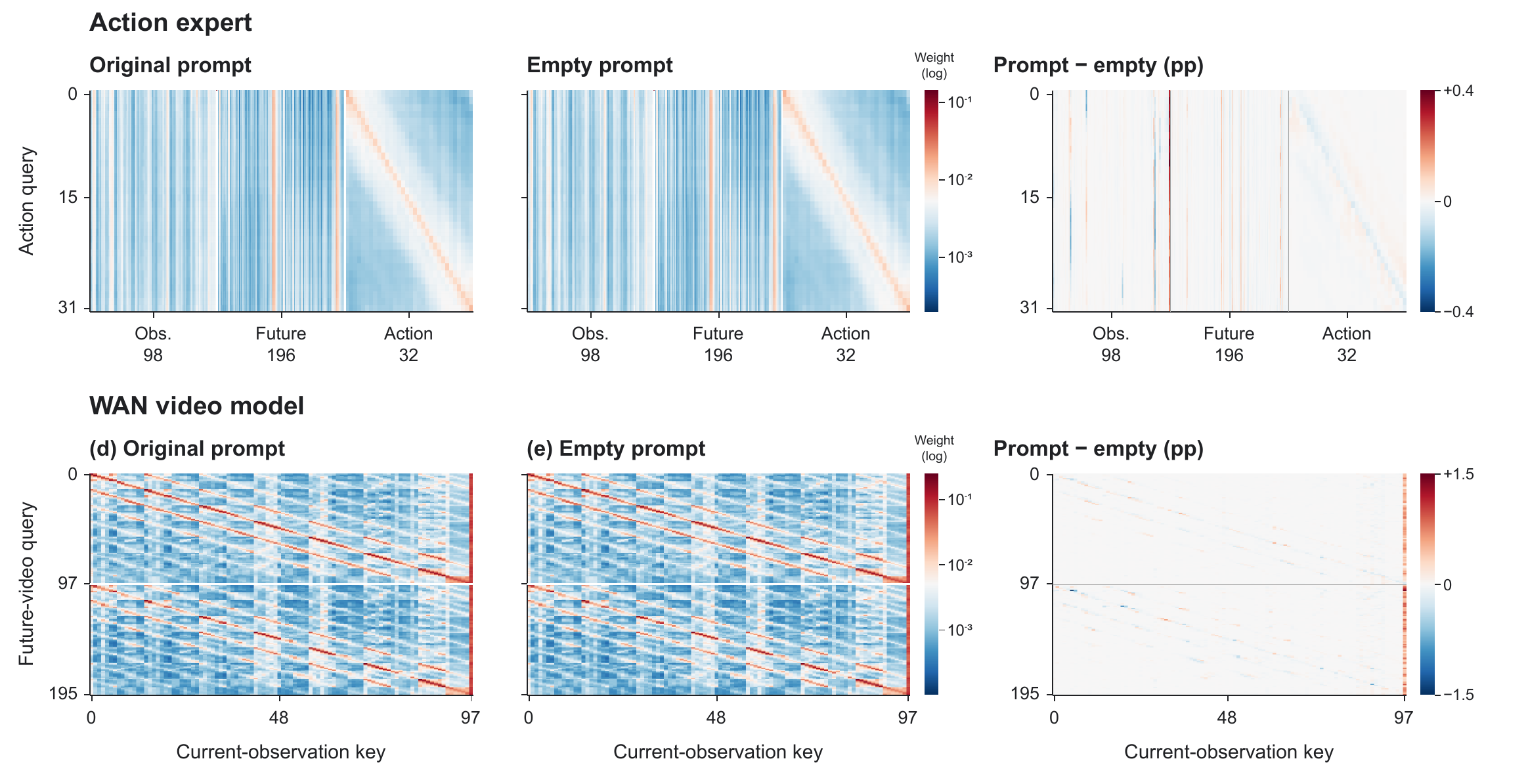}
\caption{
Attention of the frozen WAM under different instructions, with the observation and the random seed fixed. Top: attention of the action tokens. Bottom: attention of the future-video queries over the keys in the video model, under the original prompt, the empty prompt, and their difference. The two conditions produce nearly identical attention maps.
}
\label{fig:attn-prompt}
\end{figure}

Following this principle, we propose \method{} (\textbf{F}uture \textbf{A}nchored \textbf{V}erification and \textbf{O}nline \textbf{R}ecovery), which keeps the predicted frames as anchors and uses them both to verify execution and to guide recovery. Two challenges stand in the way. \textbf{(1)} Not every mismatch is a deviation: the anchor is an image produced by a generative model and may contain visual artifacts, so pixel-level differences do not necessarily imply a semantic mismatch. \textbf{(2)} The challenge is to convey the anchor to the policy and elicit the intended correction: the anchor is an image, whereas a WAM accepts only an observation and an instruction. Moreover, it leans on visual context and largely ignores the instruction: with the observation fixed, the original and the empty instruction yield nearly identical attention maps (Figure~\ref{fig:attn-prompt}). The anchor must therefore be conveyed through a corrective instruction, and the influence of that instruction must be strengthened to elicit the intended correction.

For challenge (1), we introduce the \emph{Anchor Verifier}, which compares each observation with its anchor together with the executed actions and tracks the mismatch along the chunk. It thus responds to deviations that break the task rather than to any visual difference.

For challenge (2), we introduce \emph{Anchor-Guided Recovery} to return the scene to a state the policy knows how to continue from. It hands the alarmed anchor, the current observation, and the task to a vision-language model, which issues one short instruction naming the condition to restore rather than the pixels to reproduce. The frozen WAM then executes it under strengthened instruction guidance to reach the intended future before deviation.

We evaluate \method{} on LIBERO \citep{liu2023libero} and LIBERO-Plus \citep{fei2025liberoplus}. \method{} raises the task success of the base WAM from 97.85\% to 98.10\% on LIBERO and from 72.60\% to 72.98\% on LIBERO-Plus. The Anchor Verifier separates failed from successful episodes with an AUROC of 0.992, above distance-based and learned detectors \citep{gu2025safe,xu2025faildetect}.

In summary, our main contributions are as follows:
\begin{itemize}
    \item A future-anchored view of WAM execution: the future predicted before acting serves not only as a reference for verifying execution but also as the target that recovery should restore.
    \item The Anchor Verifier, which grounds execution verification in the WAM's own predicted future and executed actions, identifying when the scene no longer supports continuation.
    \item The Anchor-Guided Recovery, which translates the alarmed anchor into a corrective instruction and strengthens instruction guidance, enabling a WAM to use its own predicted future as a recovery goal and return to a state from which the original task can continue.
\end{itemize}

%% file: section/02_related.tex
\section{Related Work}
\label{sec:related}

\subsection{World Action Models and Future Prediction}

\citationrev{WAMs integrate future prediction into robot control~\citep{shen2026wamsurvey}.} Joint video--action models couple anticipated visual changes with robot actions in a single generative process~\citep{zhu2025uwm,ye2026dreamzero,shen2025videovla,li2025uva,cen2025worldvla,zhou2026tau0wm,li2026lingbotva,ye2026gigaworld,kim2026cosmos}, whereas cascaded interfaces pass visual future states as explicit conditions to inverse dynamics models (IDMs)~\citep{du2023unipi,ko2023avdc,black2023susie,hu2024vpp,tian2025seer,wu2024gr1,liang2025video}. The role of prediction depends on the inference interface: Fast-WAM separates the benefit of video co-training from that of explicit future generation at test time~\citep{yuan2026fastwam}. Related efforts use video models as simulators or reward sources~\citep{yang2023unisim,escontrela2023viper,guo2025ctrlworld} and diagnose whether generated futures follow the commanded actions~\citep{chen2026worldecho,liu2026wav,li2026badwam}. We focus on WAMs whose predicted future states condition action generation and thus accompany the resulting actions.

In these interfaces, and in work on aligning visual features and instructions with action decoding~\citep{qiu2026agra,he2026sgwam,liu2026oawam}, the predicted future is an input to action generation and plays no further role once the actions are produced. FFDC is the exception: it retains the prediction and compares it with incoming observations to judge whether the remaining actions can still be trusted, shortening execution or replanning when they cannot~\citep{wang2026trustimagination}. \citationrev{Dense correspondence from long-range flow or all-pairs point tracking~\citep{wu2023accflow,wu2026pairformer} could also relate the two frames, but it measures motion rather than task relevance.} HarnessWAM also intervenes during WAM execution, but monitors subgoal completion rather than the prediction~\citep{gu2026harnesswam}. Here, as in adaptive-horizon methods~\citep{wang2026trustimagination,wang2026limits,nie2026pace}, the signal only decides whether to continue, and replanning starts from the deviated state.

\subsection{Execution Verification and Recovery}

Execution monitoring for learned policies draws on internal features, action consistency, and semantic progress~\citep{gu2025safe,agia2025sentinel,xu2025faildetect,du2023successvqa,duan2025aha,zhao2026verispace,liu2026checkvla,zhang2026foresight,ma2024gvl,li2026vlasafety}. \citationrev{Diagnostics for world models further expose inconsistency between imagination and execution, from rollouts that violate physical and causal consistency~\citep{chen2026harnessevalw} to actions that diverge from a correctly imagined future~\citep{li2026badwam}.} A detector, however, says only that something went wrong, not what state to return to.

Existing recovery methods obtain their target from outside the acting policy. Planner-level loops revise behavior from environmental feedback and constraint violations~\citep{huang2023inner,guo2023doremi,zhou2024code}, and test-time language correction takes the instruction from a human~\citep{liu2022sirius,liu2023olaf,shi2024yay}. Policy-level recovery for VLAs is learned from failure-and-recovery data~\citep{liu2023reflect,dai2024racer,li2024selfcorrectvla,ye2025robofac} or inferred by an auxiliary predictive component, either a dynamics model~\citep{pan2026vlacorrector} or imagined states toward which recovery is directed~\citep{shin2026b2ff,zhang2026core}. Each obtains its objective at the cost of training or a second model, and none derives it from the plan the policy was executing. \method{} takes the objective from that plan: the future the WAM predicted before acting is what recovery restores. Since a WAM accepts only an observation and an instruction, the VLM only translates this future into an instruction rather than estimating a target of its own. \citationrev{This translation step draws on VLM work in task-directed visual perception~\citep{yu2025vpt} and spatial reasoning~\citep{zhang2026geosr,xiong2026anatomy}. However, formulating an appropriate corrective instruction does not ensure that the WAM will act on it.}

\citationrev{In the WAM considered here, an IDM decodes actions from the predicted future, so the influence of a corrective instruction on execution depends largely on how it shapes future generation. Yet the WAM used here exhibits limited sensitivity to instruction changes, producing largely unchanged outputs across different instructions (Figure~\ref{fig:attn-prompt}). On the policy side, prior work makes corrections effective by training on them~\citep{liu2022sirius,liu2023olaf,shi2024yay} or by aligning visual features and instructions with action decoding~\citep{qiu2026agra,he2026sgwam,liu2026oawam}. On the generation side, classifier-free guidance strengthens a condition at inference time~\citep{ho2022classifier}, and controllable generation and editing steer diffusion models with visual conditions~\citep{tan2025ominicontrol,wu2025ifadapter,qin2026spotedit,li2025winwinlay,zhang2026corradapter,wu2024text2qr,cui2024face2qr,wu2025animateqr} or instructions~\citep{yu2026vifeedit}. \method{} strengthens the instruction on the generation side at inference time: guidance makes it shape the generated future, and the IDM turns that future into recovery actions, without retraining.}

%% file: section/03_Preliminary.tex
\section{Preliminaries}
\label{sec:method:formulation}
\input{figures/framework}

\paragraph{World action models.}
A WAM first predicts the visual states the task should pass through and then inverts them into actions \citep{du2023unipi,tian2025seer,yuan2026fastwam,wam2026tutorial}; we build on the IDM variant of Fast-WAM \citep{yuan2026fastwam}. Formally, let $o_t$ denote the observation at step $t$, and $\ell$ the task instruction. At each planning step $t_c$ the policy produces a chunk of $H$ actions $\mathbf A_c=(a_{c,0},\ldots,a_{c,H-1})$, with $a_{c,k}$ applied at step $t_c+k$. The WAM factorizes the chunk through the future $\mathbf Z_c$, a sequence of visual states over the chunk:
\begin{equation}
p(\mathbf A_c\mid o_{t_c},\ell)
=\int
\underbrace{p(\mathbf Z_c\mid o_{t_c},\ell)}_{\text{future prediction}}\;
\underbrace{p(\mathbf A_c\mid o_{t_c},\mathbf Z_c,\ell)}_{\text{inverse dynamics}}
\,\mathrm d\mathbf Z_c .
\label{eq:wam-factorization}
\end{equation}

\paragraph{Future prediction.}
The video model is a diffusion transformer \citep{peebles2023dit,wan2025wan} trained by flow matching \citep{lipman2023flow}. Let $z_t$ be the latent of $I_t$ and $\mathbf Z_c=z_{t_c+\Delta:t_c+H}$ the latents of the chunk's future frames. Flow matching learns a velocity field that transports Gaussian noise $\mathbf Z^{(1)}\sim\mathcal N(\mathbf 0,\mathbf I)$ to the future $\mathbf Z^{(0)}=\mathbf Z_c$ through the ordinary differential equation
\begin{equation}
\frac{\mathrm d\mathbf Z^{(s)}}{\mathrm ds}=\mathbf v_\theta\bigl(\mathbf Z^{(s)},s\mid z_{t_c},\ell\bigr),
\qquad s\in[0,1],
\label{eq:video-ode}
\end{equation}
conditioned on the current latent and the instruction. Rectified flow \citep{liu2023rectified} takes the straight path $\mathbf Z^{(s)}=(1-s)\,\mathbf Z_c+s\,\mathbf Z^{(1)}$, so the target velocity is simply $\mathbf Z^{(1)}-\mathbf Z_c$:
\begin{equation}
\mathbf v_\theta\bigl(\mathbf Z^{(s)},s\mid z_{t_c},\ell\bigr)\approx\mathbf Z^{(1)}-\mathbf Z_c .
\label{eq:video-flow}
\end{equation}
At inference, the future is obtained by integrating Eq.~(\ref{eq:video-ode}) backward from noise to a clean latent,
\begin{equation}
\mathbf Z^{(s_{i-1})}=\mathbf Z^{(s_i)}-(s_i-s_{i-1})\,\mathbf v_\theta\bigl(\mathbf Z^{(s_i)},s_i\mid z_{t_c},\ell\bigr),
\qquad i=T,\ldots,1,
\label{eq:future-prediction}
\end{equation}
which yields the predicted future $\widehat{\mathbf Z}_c=\mathbf Z^{(s_0)}$. Decoding $\widehat{\mathbf Z}_c$ produces $H/\Delta$ future frames $(\hat I_{c,1},\ldots,\hat I_{c,H/\Delta})$, spaced $\Delta$ control steps apart. Each frame $\hat I_{c,j}$ depicts the state expected at step $t_c+j\Delta$, with $\hat I_{c,0}=I_{t_c}$ denoting the current observation. These predicted frames serve as anchors for execution verification and recovery in Section~\ref{sec:method}.

\paragraph{Inverse dynamics.}
Given the current latent and the predicted future, the IDM infers the actions needed to realize that future \citep{yuan2026fastwam}. The action tokens are denoised with the same flow-matching procedure, conditioned on the current latent, the predicted future, and the task instruction,
\begin{equation}
\mathbf A_c\sim p(\cdot\mid z_{t_c},\widehat{\mathbf Z}_c,\ell).
\label{eq:inverse-dynamics}
\end{equation}
Predicted frames and generated actions follow a strict temporal correspondence: each group of $\Delta$ consecutive actions is meant to realize the transition between two consecutive predicted frames,
\begin{equation}
\hat I_{c,j-1}\;\xrightarrow{\;\mathbf a_{c,j}\;}\;\hat I_{c,j},\qquad
\mathbf a_{c,j}=(a_{c,(j-1)\Delta},\ldots,a_{c,j\Delta-1}),\qquad
j=1,\ldots,H/\Delta,
\label{eq:frame-action}
\end{equation}

with $\hat I_{c,0}=I_{t_c}$ for the first action.
The action $\mathbf a_{c,j}$ is executed from step $t_c+(j-1)\Delta$ through step $t_c+j\Delta-1$, so that $\hat I_{c,j}$ describes the state these actions are meant to reach at step $t_c+j\Delta$. The observation $I_{t_c+j\Delta}$ can then be compared with $\hat I_{c,j}$ to check whether execution has reached it.

%% file: figures/framework.tex
\begin{figure}[t]
\centering
\includegraphics[width=\textwidth]{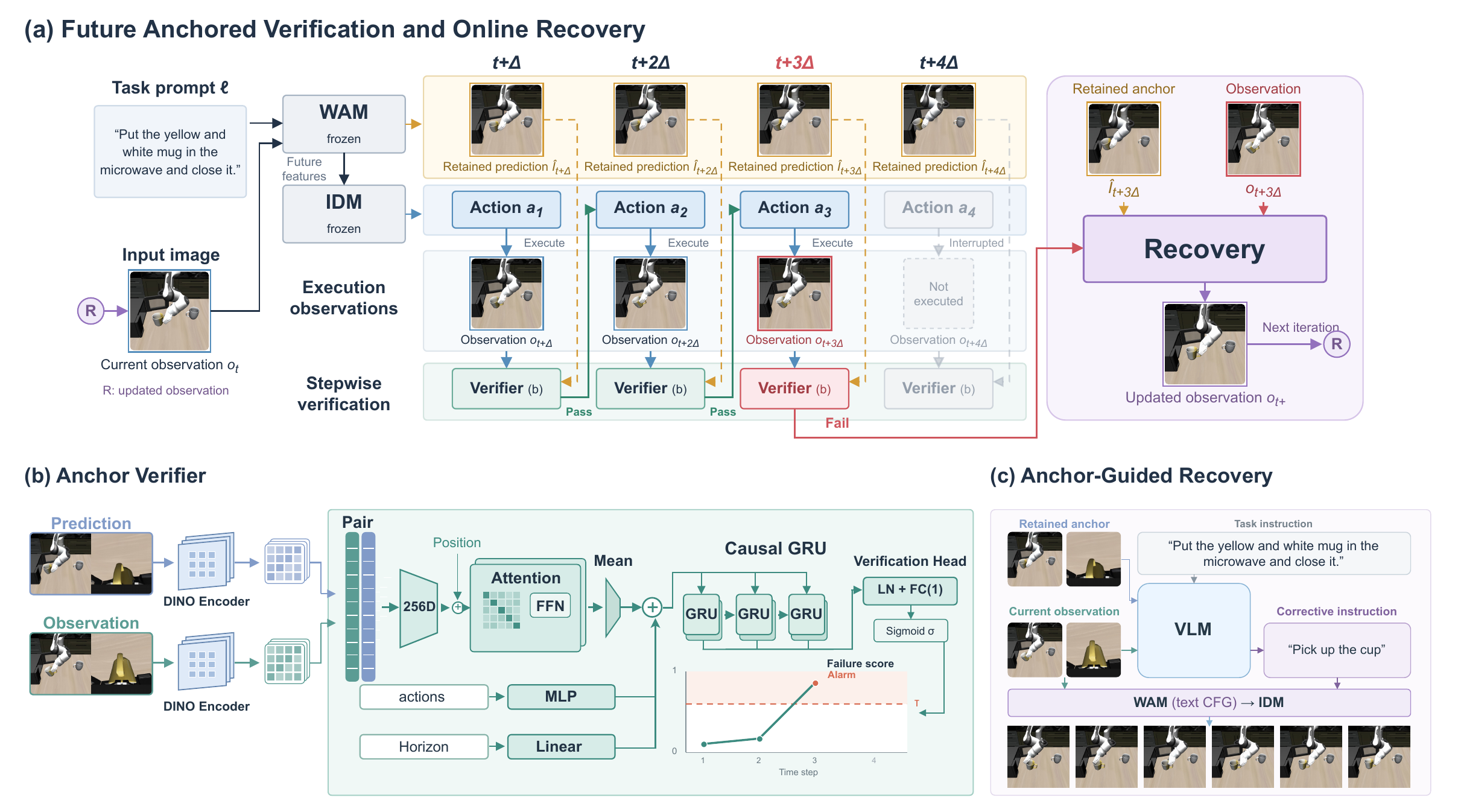}
\caption{Overview of \method{}. \textbf{(a)} The frozen WAM predicts future frames and the IDM decodes an action chunk; the predicted frames are retained as anchors. Every $\Delta$ steps, the verifier compares the observation with its anchor: a pass continues execution, and the failure interrupts the remaining actions and triggers recovery. \textbf{(b)} The Anchor Verifier encodes the prediction--observation pair with DINO, fuses it with action and horizon embeddings, and outputs a failure score through a causal GRU; an alarm is raised when the score exceeds $\tau$. \textbf{(c)} Anchor-Guided Recovery: a VLM reads the alarmed anchor, the current observation, and the task, and issues a corrective instruction.}

\label{fig:framework}
\end{figure}

%% file: section/04_method.tex
\section{Method}
\label{sec:method}
\paragraph{Overview.}

As observed in Section~\ref{sec:intro}, the future predicted before acting depicts the states the chunk was meant to reach, and these states lie where the policy is reliable. \method{} therefore keeps the predicted frames as \emph{anchors} and uses them twice (Figure~\ref{fig:framework}). An Anchor Verifier checks whether the current observation has deviated too far from its anchor, and Anchor-Guided Recovery uses that anchor to guide the robot back on course.

\subsection{Anchor Verifier}
\label{sec:method:compare}

Whether an observation still matches its anchor $\hat I_{c,j}$ need be judged semantically and over time. A fixed image distance~\citep{wang2026trustimagination} captures neither, so we learn the verifier from where the mismatch lies, how much motion the actions $\mathbf a_{c,j}$ command, and how it evolves along the chunk.

\paragraph{Verifier architecture.}
As shown in Figure~\ref{fig:framework}(b), at anchor $j$ of chunk $c$ a frozen DINOv2 encoder \citep{oquab2024dinov2} tokenizes the observation $I_{t_c+j\Delta}$ and the anchor $\hat I_{c,j}$ into $O_{c,j}$ and $\hat O_{c,j}$. Together with the executed actions $\mathbf a_{c,j}$ and the anchor horizon $p_j=j\Delta/H$, they form the input
\begin{equation}
X_{c,j}=\bigl[\,O_{c,j},\;\hat O_{c,j},\;\mathbf a_{c,j},\;p_j\,\bigr].
\label{eq:verifier-input}
\end{equation}
Observation and anchor tokens are paired position by position and fused by a spatial Transformer into a visual summary, to which embeddings of the actions and of the position are added, giving the anchor feature $x_{c,j}$. A GRU \citep{cho2014gru} carries these features across the anchors of the chunk and a linear head $g$ produces the logit $u_{c,j}$ and the deviation score $s_{c,j}$
\begin{equation}
u_{c,j}=g\bigl(\operatorname{GRU}(x_{c,1},\ldots,x_{c,j})\bigr),
\qquad
s_{c,j}=\sigma(u_{c,j})\in[0,1].
\label{eq:score}
\end{equation}
An alarm is raised when $s_{c,j}$ exceeds a threshold $\tau$ fixed before evaluation.

\paragraph{Training data.}
To train the Anchor Verifier, we construct two datasets with different levels of supervision. The first is a \emph{locally labeled} set with a deviation label at every anchor. From the same simulator state, we execute a clean and a perturbed version of the planned actions and label each anchor by whether the perturbed execution has departed from the clean one. We use two perturbations, late Gaussian noise added to the arm commands in the latter half of the chunk and tail scaling that multiplies these commands by a fixed factor. The second is an \emph{episode-labeled} set that carries only a terminal success or failure label per episode; it contains natural rollouts of the policy together with the demonstrations from its training set, which all count as successes.

\paragraph{Training strategy.}
We initialize the verifier using local supervision and then adapt it to natural rollouts using terminal outcomes. Since a failed rollout may contain many anchors that were correctly reached, we use multiple-instance learning to supervise episode-level predictions. Let $u_{i,c,j}$ denote the verifier logit at anchor $(c,j)$ of episode $i$, $\mathcal V_i$ the set of its valid anchors, and $m_i=\max_{(c,j)\in\mathcal V_i}u_{i,c,j}$ the episode logit. For the failed and successful episode sets
$\mathcal F$ and $\mathcal S$ in each training batch, we minimize
\begin{align}
L_F &=
\frac{1}{|\mathcal F|}
\sum_{i\in\mathcal F}\operatorname{sp}(-m_i), \\
L_S &=
\frac{1}{|\mathcal S|}
\sum_{i\in\mathcal S}
\left[
\frac{1}{2}\operatorname{sp}(m_i)
+
\frac{1}{2|\mathcal V_i|}
\sum_{(c,j)\in\mathcal V_i}
\operatorname{sp}(u_{i,c,j})
\right], \\
L_{\mathrm{ep}} &= \frac{1}{2}(L_F+L_S),
\qquad
\operatorname{sp}(x)=\log(1+e^x).
\label{eq:episode-loss}
\end{align}

The failure term encourages at least one high score anchor without assigning positive targets throughout a failed episode. The success term suppresses both the strongest and the average score of successful episodes, so that alarms stay rare while execution is on track.

\subsection{Anchor-Guided Recovery}
\label{sec:method:recovery}

\begin{figure}[t]
\centering
\includegraphics[width=\textwidth]{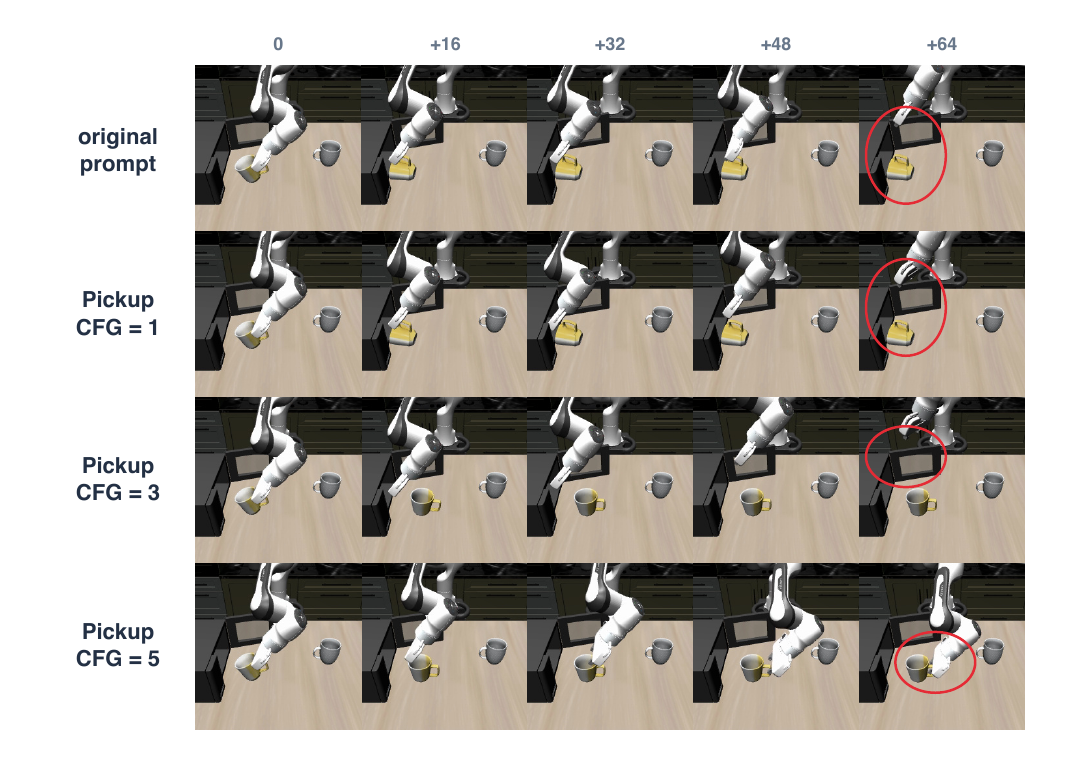}
\caption{
Effect of the guidance scale on recovery predictions.
All branches start from the same trigger state and the corrective instruction ``pick up the cup''.
}
\label{fig:cfg-ablation}
\end{figure}

At the alarm, the controller discards the unexecuted actions of the chunk and keeps the alarmed anchor $I^{\mathrm{anc}}$. The WAM is conditioned on a current observation and an instruction, so the anchor needs to reach it through language. Specifically, we use a VLM $\mathcal M$ to generate a corrective instruction for transitioning from the current observation $I_{t^\star}$ to the anchored state $I^{\mathrm{anc}}$.
\begin{equation}
 \ell_{\mathrm{corr}}=\mathcal M(I^{\mathrm{anc}},I_{t^\star},\ell).
 \label{eq:recovery-instruction}
\end{equation}
The anchor specifies the nominal state that execution was expected to reach. Expressing this target as a short instruction makes the correction depend on the corresponding object-level state change, such as grasping an object, rather than on exact visual agreement with the predicted frame. If the VLM judges that no correction is needed, it returns the original task, $\ell_{\mathrm{corr}}=\ell$. The WAM then generates futures and actions from the current observation and $\ell_{\mathrm{corr}}$, exactly as in normal execution.

Recovery starts from a state outside the policy's training distribution under an instruction it never saw, and Figure~\ref{fig:attn-prompt} shows that the instruction has little influence in the WAM. We apply classifier-free guidance (CFG) \citep{ho2022classifier} during recovery to strengthen the effect of $\ell_{\mathrm{corr}}$ on the predicted future:
\begin{equation}
\hat{\mathbf v}_\theta\bigl(\mathbf Z^{(s)},s\mid z_{t^\star},\ell_{\mathrm{corr}};\gamma\bigr)
=\mathbf v_\theta(\cdot\mid z_{t^\star},\varnothing)
+\gamma\bigl[\mathbf v_\theta(\cdot\mid z_{t^\star},\ell_{\mathrm{corr}})-\mathbf v_\theta(\cdot\mid z_{t^\star},\varnothing)\bigr],
\label{eq:cfg}
\end{equation}
where $\varnothing$ denotes the empty instruction and $\gamma$ the guidance scale (Figure~\ref{fig:cfg-ablation}). The IDM then decodes the guided future into recovery actions.

At the end of the window, the VLM compares the current observation with the anchor once more: if it judges that the anchored condition is restored, control returns to the policy.

%% file: section/05_experiments.tex
\section{Experiments}
\label{sec:exp}
\begin{table*}[t]
    \centering
    \small
    \setlength{\tabcolsep}{2.6pt}
    \renewcommand{\arraystretch}{1.1}
    \caption{
        Task success (\%, $\uparrow$) on LIBERO and LIBERO-Plus, and the LIBERO failure rate (\%, $\downarrow$) with its relative change from the base WAM. Bold marks the best entry in each column; the last row gives the gain of \method{} over the base WAM in percentage points.
    }
    \label{tab:overall}
    \begin{tabular}{@{}lcccccccc@{}}
        \toprule
        & \multicolumn{7}{c}{LIBERO} & LIBERO-Plus \\
        \cmidrule(lr){2-8}\cmidrule(lr){9-9}
        & \multicolumn{5}{c}{Success (\%, $\uparrow$)} & \multicolumn{2}{c}{Failure (\%, $\downarrow$)} & Success ($\uparrow$) \\
        \cmidrule(lr){2-6}\cmidrule(lr){7-8}\cmidrule(lr){9-9}
        Method & Spatial & Object & Goal & Long & Avg. & Rate & Rel. & Avg. \\
        \midrule
        Fast-WAM-IDM
        & 98.20 & 98.90 & 98.30 & 96.00 & 97.85 & 2.15 & -- & 72.60 \\
        \midrule
        + Verifier, direct replanning
        & \textbf{98.30} & 99.20 & 98.20 & 96.00 & 97.92 & 2.08 & $-$3.3\% & 72.60 \\
        + Verifier, correction w/o anchor
        & \textbf{98.30} & 99.30 & \textbf{98.40} & 96.10 & 98.02 & 1.98 & $-$7.9\% & 72.53 \\
        \rowcolor{gray!15}
        \method{} (Anchor-Guided Recovery)
        & \textbf{98.30} & \textbf{99.40} & \textbf{98.40} & \textbf{96.30} & \textbf{98.10} & \textbf{1.90} & $\mathbf{-}$\textbf{11.6\%} & \textbf{72.98} \\
        \midrule
        \textcolor{black!60}{Gain over base (pp)}
        & \textcolor{black!60}{$+$0.10} & \textcolor{black!60}{$+$0.50} & \textcolor{black!60}{$+$0.10} & \textcolor{black!60}{$+$0.30} & \textcolor{black!60}{$+$0.25} & \textcolor{black!60}{$-$0.25} & & \textcolor{black!60}{$+$0.37} \\
        \bottomrule
    \end{tabular}
\end{table*}

\subsection{Experimental Setup}
\label{sec:exp:setup}

\paragraph{Benchmarks and evaluation protocols.}
We evaluate \method{} on LIBERO \citep{liu2023libero} and LIBERO-Plus \citep{fei2025liberoplus}. LIBERO comprises four suites of ten tasks each: Spatial, Object, Goal, and Long (LIBERO-10). LIBERO-Plus uses the LIBERO tasks while varying the environment, task configuration, and observation distribution beyond the standard setting. We run 100 rollouts per task on LIBERO and a total of 1698 rollouts on LIBERO-Plus.

\paragraph{Implementation details.}
We instantiate \method{} on the IDM variant of Fast-WAM \citep{yuan2026fastwam}. The Anchor Verifier is trained on one NVIDIA A5000 GPU (24\,GB). We use Qwen3-VL-2B-Instruct \citep{qwen3vl} as the VLM in Anchor-Guided Recovery. All evaluations run on four A5000 GPUs.

\paragraph{Evaluation metrics.}
The primary metric is native task success, determined by the environment's own success condition and computed over the complete evaluation cohort, including episodes in which no intervention occurs. It thus credits successful recoveries and penalizes interventions that disrupt otherwise successful execution. The Anchor Verifier and Anchor-Guided Recovery are additionally evaluated in isolation in Section~\ref{sec:exp:ablation}.

\subsection{Main Results}
\label{sec:exp:overall}
\label{sec:results:overall}

\paragraph{Results on LIBERO and LIBERO-Plus.}
Table~\ref{tab:overall} reports native task success. \method{} raises the average success of the base WAM from 97.85\% to 98.10\% on LIBERO, cutting the failure rate by 11.6\% relative, and from 72.60\% to 72.98\% on LIBERO-Plus. Both components are needed. From the same alarms, direct replanning under the original instruction removes only 3.3\% of the base failures on LIBERO and none on LIBERO-Plus. Language correction without the anchor removes 7.9\% on LIBERO but slightly hurts on LIBERO-Plus (72.53\%); supplying the anchor accounts for the rest. The gain is largest on Object and Long. Object tasks share one scene and differ only in the target object, so the policy relies more on the instruction and a corrective instruction moves it more; on Long, an early deviation has the most time to compound.

\paragraph{Qualitative results.}
\label{sec:results:failure}
\label{sec:analysis:recovery}

\begin{figure}[t]
\centering
\includegraphics[width=\textwidth]{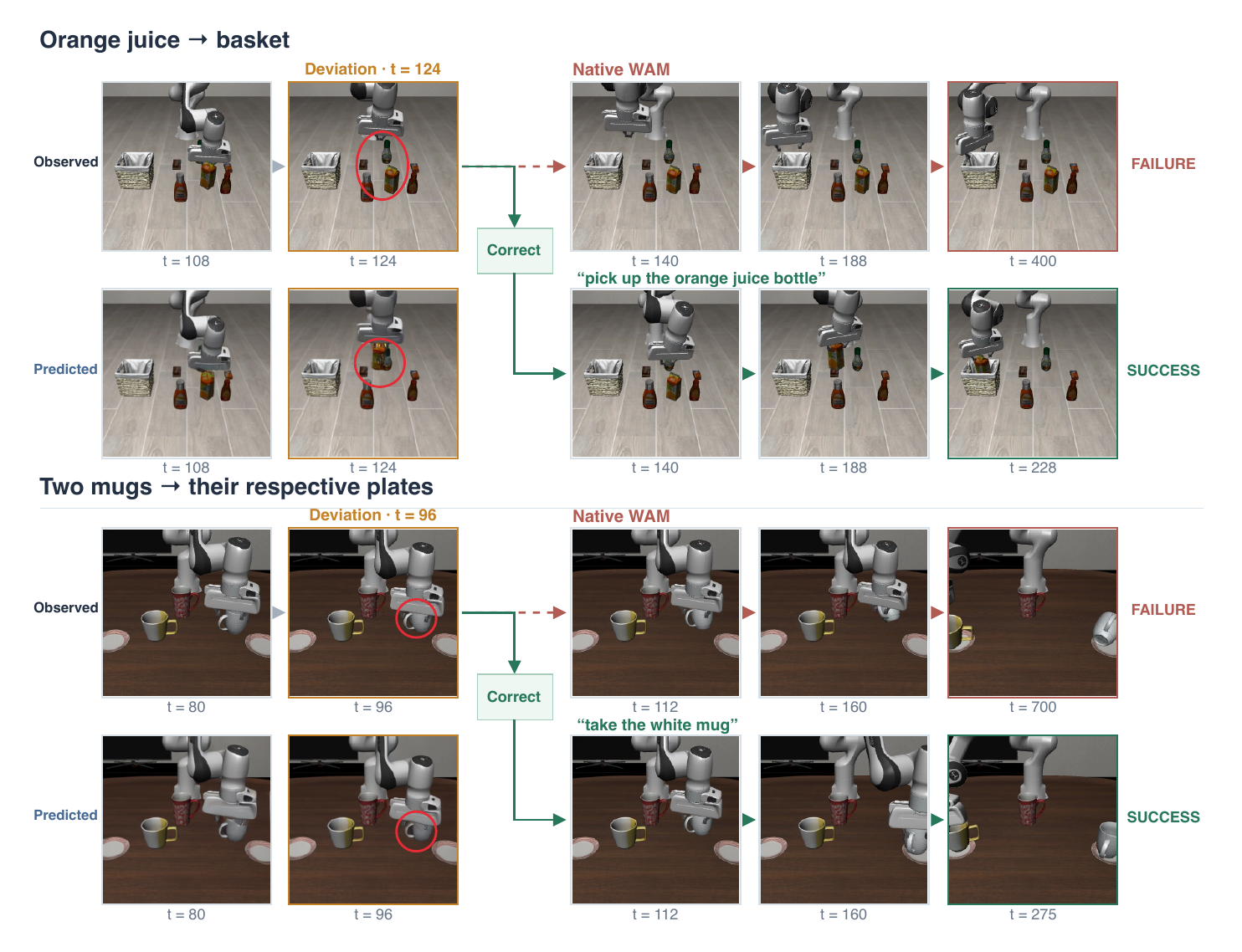}
\caption{
Qualitative examples of Anchor-Guided Recovery on LIBERO.
In each case, the top row shows the observed execution and the bottom row
the WAM's retained prediction up to the alarm (orange frame);
red circles mark where the observation departs from its anchor.
Right of the alarm, native continuation under the original instruction fails (top),
while the corrective instruction issued by the VLM (green) restores the anchored
condition and the task succeeds (bottom).
Both branches start from the same simulator state at the alarm.
}
\label{fig:paired-cases}
\end{figure}

Figure~\ref{fig:paired-cases} shows two cases that fail under native continuation and are rescued by \method{}. In both, the Anchor Verifier detects the deviated state as soon as the observation stops matching its anchor: a bottle that the plan expects in the gripper but is not, and a mug that has left its intended path. Anchor-Guided Recovery reads the anchor to identify what was lost and issues a short instruction that restores it, ``pick up the orange juice bottle'' and ``take the white mug''. The scene returns to the anchored state, the original plan applies again, and both tasks complete. Native continuation, by contrast, keeps executing the remaining actions as if nothing had happened, and both tasks fail once the budget runs out.

\subsection{Ablation Studies}
\label{sec:exp:ablation}

\label{sec:exp:analysis}

\paragraph{Anchor Verifier.}
\label{sec:exp:verification}
\label{sec:results:verification}

We compare the Anchor Verifier with DINO adaptations of SAFE-MLP and SAFE-LSTM~\citep{gu2025safe}, and FAIL-Detect logpZO~\citep{xu2025faildetect}, as well as DINO feature distance based on DINOv2~\citep{oquab2024dinov2} and PSNR~\citep{huynhthu2008psnr}. All six methods score the same saved base-WAM test episodes, whose initial states are disjoint from those used to train, validate, and calibrate the learned detectors.

\begin{figure}[t]
\centering
\begin{minipage}{\textwidth}
\centering\small\setlength{\tabcolsep}{6pt}
    \captionof{table}{Detector comparison on LIBERO.}
    \label{tab:verification}
    \begin{tabular}{@{}lccc@{}}
        \toprule
        Detector & AUROC $\uparrow$ & AUPRC $\uparrow$ & Coverage $\uparrow$ \\
        \midrule
        SAFE-MLP (DINO)~\citep{gu2025safe} & 0.935 & 0.664 & 67.6\% \\
        SAFE-LSTM (DINO)~\citep{gu2025safe} & 0.712 & 0.152 & 38.2\% \\
        FAIL-Detect logpZO (DINO)~\citep{xu2025faildetect} & 0.951 & 0.512 & 73.5\% \\
        DINO feature distance~\citep{oquab2024dinov2} & 0.944 & 0.401 & 73.5\% \\
        PSNR~\citep{huynhthu2008psnr} & 0.956 & 0.472 & 67.6\% \\
        \midrule
        \textbf{Anchor Verifier (ours)} & 0.992 & 0.781 & 97.1\% \\
        \bottomrule
    \end{tabular}
\end{minipage}
\vspace{6pt}

\includegraphics[width=0.82\textwidth]{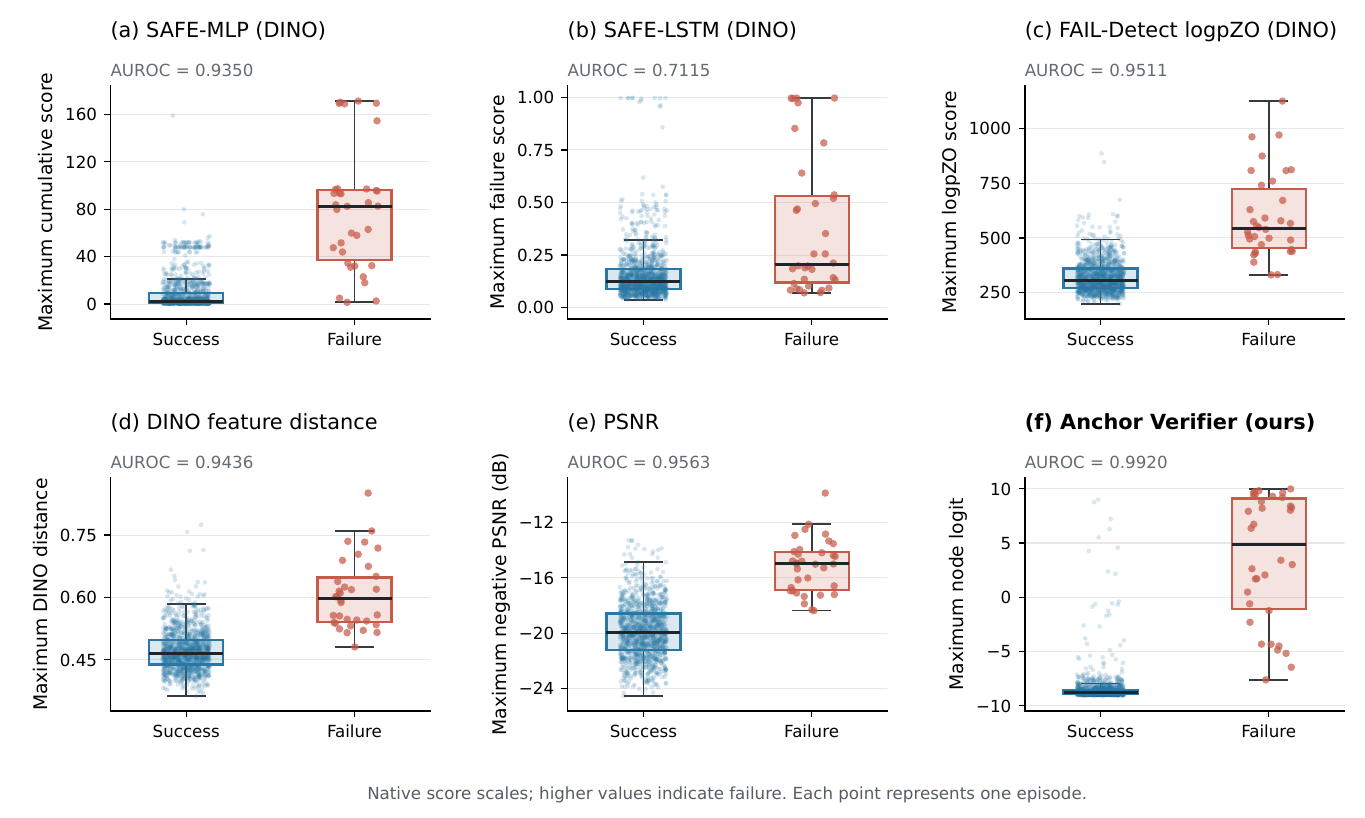}
\caption{Distribution of the maximum episode score for successful and failed LIBERO episodes under each detector, on its native scale (higher indicates failure); each point is one episode.}
\label{fig:verifier-libero}
\end{figure}

Each episode is scored by its maximum anomaly score over the anchors of complete chunks. The baselines split into learned detectors that do not see the anchor and fixed comparisons with it. SAFE-MLP and SAFE-LSTM \citep{gu2025safe} learn a failure classifier from terminal labels on the same DINO features of the observation, per anchor and over the anchor sequence respectively. FAIL-Detect logpZO \citep{xu2025faildetect} fits a flow model to the observation features of successful episodes and scores an episode by how unlikely its features are under that model. DINO distance averages Euclidean differences between corresponding observed and anchor tokens, and PSNR uses the mean squared RGB error between the observation and the anchor across both cameras, negated so that higher scores indicate greater discrepancy; neither is trained. Table~\ref{tab:verification} reports episode-level AUROC and AUPRC with task failure as the positive class, and failure coverage at each detector's own threshold, calibrated on successful episodes at a 5\% alarm rate and fixed before testing.

\ifdefined\arxivversion\enlargethispage{-\baselineskip}\fi %
The Anchor Verifier leads on all three metrics. The fixed comparisons rank episodes well (AUROC 0.94--0.96) but have low AUPRC and coverage, so a raw distance to the anchor cannot tell a lost grasp from a harmless mismatch. The learned detectors without the anchor do no better, and SAFE-LSTM collapses. Learning the comparison against the anchor combines the two and raises AUPRC by 0.12 and coverage by 24 points. Figure~\ref{fig:verifier-libero} shows why. Under every baseline, the scores of successful episodes spread over the same range as those of failures. Under the Anchor Verifier, the successes collapse into a tight band at the bottom of the scale and the failures rise well above it, so one threshold separates them, yielding high coverage at a low alarm rate.

\paragraph{Anchor-Guided Recovery.}
\label{sec:exp:guidance}

We ablate the guidance scale of Anchor-Guided Recovery: Figure~\ref{fig:cfg-ablation} fixes the state and the corrective instruction ``pick up the cup'' and varies only $\gamma$ in Eq.~(\ref{eq:cfg}). At $\gamma=1$, which reduces to plain conditional sampling, the predicted future is nearly identical to the continuation under the original task: the correction is ignored. At $\gamma=3$ the prediction begins to depart from that continuation, but it does not yet realize the instruction, as the arm leaves the original path without reaching a grasp; the correction is registered but not followed. At $\gamma=5$ the predicted future shows the arm grasping the cup, and the IDM turns this future into the corresponding recovery actions. The guidance scale thus sets how far the corrective instruction overrides the policy's habitual continuation, and it must be large enough for the instruction to prevail.

%% file: section/07_conclusion.tex
\section{Conclusion}
\label{sec:conclusion}

World action models plan by predicting the future, yet once execution drifts from that plan they have no way to notice or to recover. \method{} adds this capability to a frozen WAM by reusing its own prediction. The Anchor Verifier compares observations with their anchors to detect deviations that break the task, and Anchor-Guided Recovery turns the alarmed anchor into a corrective instruction that steers the policy back to a state it can continue from. On LIBERO and LIBERO-Plus it reduces failures and improves end-to-end success without modifying the policy.

%% file: section/07_statements.tex
\ifdefined\arxivversion\else
\clearpage
\subsection*{AI use statement}
In this work, we used generative AI tools to edit the paper for readability, to suggest the structure of sections, and to format references. We have reviewed all AI-assisted work: every edited passage was checked by the authors against the underlying results, and all claims, numbers, and citations were verified manually. We take responsibility for the final content of this work, including text, claims, or artifacts produced with the aid of generative AI.

\subsection*{Reproducibility statement}
The appendix, provided as supplementary material, contains the details needed to reproduce our results: Appendix~\ref{app:controller} gives the complete controller as pseudocode, Appendix~\ref{app:impl} lists the default settings and the training procedure of the Anchor Verifier, and Appendix~\ref{app:prompts} reproduces the exact prompt given to the VLM together with its decoding settings. All experiments use the public LIBERO and LIBERO-Plus benchmarks and the released Fast-WAM checkpoints. 
\fi

%% file: section/08_appendix.tex
\section{Complete Controller}
\label{app:controller}

Algorithm~\ref{alg:favor} gives the complete \method{} controller used in all experiments.

\begin{algorithm}[H]
\caption{\method{}}
\label{alg:favor}
\begin{algorithmic}[1]

\Require Frozen WAM--IDM policy $\pi$, VLM $\mathcal{M}$,
         verifier $\mathcal{V}$, task instruction $\ell$;
         chunk horizon $R$, verification interval $\Delta$,
         correction horizon $B$, threshold $\tau$

\State Observe $o_0$; $t\gets0$
\State $\ell_{\mathrm{act}}\gets\ell$
\State $\mathrm{intervened}\gets\mathrm{false}$

\While{task not successful}

    \State Predict future states and generate an action chunk
           from $(o_t,\ell_{\mathrm{act}})$

    \For{$k=1,\ldots,R$}

        \State Execute $a_k$ and observe $o_{t+k}$

        \If{not $\mathrm{intervened}$ and $k\bmod\Delta=0$}

            \State Compare $o_{t+k}$ with its corresponding
                   predicted state using $\mathcal{V}$

            \If{$s_k>\tau$}

                \State Select the corresponding predicted state
                       as anchor $I^{\mathrm{anc}}$

                \State $\ell_{\mathrm{corr}}
                       \gets
                       \mathcal{M}(o_{t+k},I^{\mathrm{anc}},\ell)$

                \Repeat
                    \State Execute $\ell_{\mathrm{corr}}$ with guidance scale $\gamma$ for $B$ steps
                    \State $\ell_{\mathrm{corr}}\gets\mathcal{M}(o_{t},I^{\mathrm{anc}},\ell)$
                \Until{$\mathcal{M}$ reports that the anchor is restored, or the budget is exhausted}

                \State $\ell_{\mathrm{act}}\gets\ell$
                \State $\mathrm{intervened}\gets\mathrm{true}$

                \State \textbf{break}
                \Comment{Discard the remaining action suffix}

            \EndIf
        \EndIf

        \If{task successful}
            \State \Return
        \EndIf

    \EndFor

    \State Update the current observation and replan

\EndWhile

\end{algorithmic}
\end{algorithm}

\section{Implementation Details}
\label{app:impl}

\paragraph{Default settings.}
Table~\ref{tab:appendix-settings} lists the settings that the main text refers to symbolically.

\begin{table}[H]
\centering\small
\caption{Default settings of \method{} in our experiments.}
\label{tab:appendix-settings}
\begin{tabularx}{\linewidth}{@{}ll>{\raggedright\arraybackslash}X@{}}
\toprule
Symbol & Meaning & Value \\
\midrule
$H$ & Predicted actions per chunk & 16 \\
$R$ & Executed actions per chunk before replanning & 16 \\
$\Delta$ & Actions between consecutive anchors & 4 \\
$\tau$ & Alarm threshold of the fixed operating point & 0.5 (LIBERO), 0.9 (LIBERO-Plus) \\
$B$ & Maximum length of the correction window (actions) & 64 \\
$\gamma$ & Guidance scale during recovery & 5 \\
$\lambda$, $\eta$ & Failure weight and auxiliary weight in $L_{\mathrm{local}}$ & $2$, $0.1$ \\
$T$ & Action budget per suite & $400$~(Spatial/Object/Goal), $700$~(Long) \\
\bottomrule
\end{tabularx}
\end{table}

\paragraph{Training data.}
We construct paired execution snippets and complete policy rollouts.
For each paired snippet, we save a simulator state and use the frozen
policy to generate a 16-step action sequence and its imagined visual
trajectory. We restore the same state for clean and perturbed executions,
sharing the original imagined reference. Perturbations either add
Gaussian noise to the arm commands or scale their magnitudes over the
last eight steps, while leaving gripper commands unchanged. We record
both branches' observations and physical states to derive local
deviation labels from their object and end-effector displacements.

For complete rollouts, we execute the frozen policy until success or
the action budget is exhausted. We record the task instruction,
initial-state identifier and random seed, imagined frames and planned
actions at each replanning point, executed actions, synchronized
external and wrist-camera observations, robot and object states, and
the terminal task outcome. Observations are aligned with imagined
frames after actions 4, 8, 12, and 16 of each chunk; incomplete trailing
chunks are excluded. Paired snippets provide local deviation labels, whereas complete rollouts provide terminal outcomes.

\paragraph{Episode-level training.}
The episode-level stage warm-starts from a checkpoint previously trained with paired data and localized labels and optimizes $L_{\mathrm{ep}}$ in Eq.~(\ref{eq:episode-loss}); the episode maximum spans the valid anchors of all complete chunks, and padding contributes to no loss term. Each epoch visits every training success once, batching up to 7 successes with one cyclically sampled failure, so that all training failures are covered. We use AdamW with a learning rate of $10^{-5}$, weight decay of $10^{-4}$, and gradient norm clipping at $1.0$ for $5$ epochs. 

\paragraph{Locally supervised refinement.}
The episode maximum may become high only after an error has become irreversible. To study temporal specificity, a training variant supplements terminal-outcome supervision with reviewed anchor labels:
\begin{equation}
 L_{\mathrm{local}}=\frac{\lambda L_F+L_S}{\lambda+1}+\eta L_{\mathrm{ann}}.
 \label{eq:training}
\end{equation}
Here $\lambda$ weights failed episodes and $\eta$ weights the auxiliary local-label loss. This refinement changes training but not the inference architecture.

Specifically, $L_{\mathrm{ann}}=0.5L_{+}+0.5L_{\mathrm{event}}+L_{-}+0.2L_{\mathrm{rank}}+0.1L_{\mathrm{hard}}$. The per-anchor losses $L_{+}$ and $L_{-}$ are binary cross-entropy on reviewed positive and negative anchors, averaged within each annotated episode and then across episodes. For the event term $L_{\mathrm{event}}$, events are contiguous runs of positive anchors within a chunk, pooled from logits $u$ by $\operatorname{pool}(u)=0.5\log\operatorname{mean}\exp(2u)$. The ranking term averages $\operatorname{sp}(0.5-\operatorname{pool}(u_+)+\operatorname{pool}(u_-))$ against reviewed-negative bags from the same task, including the same episode. The hard-negative term averages the negative-target binary cross-entropy over the four highest reviewed-negative logits of each episode, or over all of them if fewer exist. Successful episodes are excluded from this term, and empty terms contribute zero.

\section{VLM Prompts for Diagnosis and Recovery Instructions}
\label{app:prompts}

The deployed prompt substitutes the original task instruction below. Each image combines the external and wrist views; the predicted anchor precedes the current observation.
\begin{quote}
\small\ttfamily\raggedright
Task: \{task\_instruction\}\par\medskip
Image 1 is the nominal state expected at this moment.\par
Image 2 is the current observed state.\par
Give one short robot command that moves the current state toward the nominal state. If they already match, reply "continue the task". Reply with the command only.
\end{quote}
Qwen3-VL-2B-Instruct runs in bfloat16 with deterministic decoding, an image longest edge of 384 pixels, and at most 64 new tokens. Raw responses and input images are retained; no subsequent ground-truth outcome enters the prompt.

\paragraph{Empty-instruction control.}
For the language analysis in Section~\ref{sec:method:recovery}, the literal empty string bypasses the usual instruction template. Its cached text encoding retains one valid special token, and padding and the original attention interface remain unchanged; the text branch is therefore not removed. This control differs from setting the video guidance scale to zero, which affects video denoising while the action stage still receives the task text.